\documentclass[conference]{IEEEtran}
\IEEEoverridecommandlockouts

\usepackage[hidelinks]{hyperref}
\usepackage[cmex10]{amsmath}
\usepackage{amssymb,amsfonts}
\usepackage{dblfloatfix}

\usepackage[ruled,vlined]{algorithm2e}
\usepackage{graphicx}
\graphicspath{{Figures/PDF/}{Figures/PNG/}}

\usepackage{xcolor}
\definecolor{agrcolor}{RGB}{142, 210, 107}
\definecolor{clicolor}{RGB}{216, 229, 244}
\definecolor{urbcolor}{RGB}{230, 195, 231}
\definecolor{forcolor}{RGB}{231, 93, 64}
\definecolor{satcolor}{RGB}{241, 184, 53}
\definecolor{datcolor}{RGB}{255, 251, 205}
\definecolor{dbcolor}{RGB}{242, 219, 207}
\definecolor{mapcolor}{RGB}{214, 239, 202}

\usepackage{tcolorbox}
\usepackage{booktabs}
\usepackage{siunitx}
\usepackage[numbers,compress]{natbib}
\usepackage{texnames}
\usepackage{bm,bbm}
\usepackage{orcidlink}
\usepackage{multirow}

\begin{document}

\title{\uppercase{Multi-Agent Geospatial Copilots for Remote Sensing Workflows}
\thanks{$^*$Equal contribution $\diamond$Work while at Microsoft; currently new affiliations.}
}

\author{Chaehong Lee$^{\dagger*\diamond}$\orcidlink{0009-0006-3319-3920}, Varatheepan~Paramanayakam$^{\S*}$\orcidlink{0009-0000-1982-9693}, Andreas~Karatzas$^{\S}$\orcidlink{0000-0001-6804-135X}, Yanan Jian$^{\dagger}$\orcidlink{0009-0000-7056-3616}, Michael Fore$^{\dagger\diamond}$\orcidlink{0009-0001-7721-3704}, \\ \IEEEauthorblockN{Heming Liao$^{\dagger}$\orcidlink{0009-0000-7056-3616}, Fuxun Yu$^{\dagger}$\orcidlink{0000-0002-4880-6658}, Ruopu Li$^{\P}$\orcidlink{0000-0003-3500-0273}, Iraklis Anagnostopoulos$^{\S}$\orcidlink{0000-0003-0985-3045}, Dimitrios Stamoulis$^{\ddagger*}$\orcidlink{0000-0003-1682-9350}}
\IEEEauthorblockA{$^{\ddagger}$Department of Electrical and Computer Engineering, The University of Texas at Austin, Austin, TX, USA}
\IEEEauthorblockA{$^{\S}$School of Electrical, Computer and Biomedical Engineering, Southern Illinois University, Carbondale, IL, USA}
\IEEEauthorblockA{$^{\P}$School of Earth Systems and Sustainability, Southern Illinois University, Carbondale, IL, USA}
\IEEEauthorblockA{$^\dagger$Microsoft Corporation, Redmond, WA, USA -- Email:  dstamoulis@utexas.edu}
}

\maketitle
\begin{abstract} 
We present GeoLLM-Squad, a geospatial Copilot that introduces the novel \textit{multi-agent} paradigm to remote sensing (RS) workflows. Unlike existing single-agent approaches that rely on monolithic large language models (LLM), GeoLLM-Squad \textit{separates agentic orchestration from geospatial task-solving}, by delegating RS tasks to specialized sub-agents. Built on the open-source AutoGen and GeoLLM-Engine frameworks, our work enables the modular integration of diverse applications, spanning urban monitoring, forestry protection, climate analysis, and agriculture studies. Our results demonstrate that while single-agent systems struggle to scale with increasing RS task complexity, GeoLLM-Squad maintains robust performance, achieving a 17\% improvement in agentic correctness over state-of-the-art baselines. Our findings highlight the potential of multi-agent AI in advancing RS workflows.
\end{abstract}


\section{Introduction}

Remote sensing (RS) workflows are inherently complex, requiring diverse datasets, tools, and analytical methods, all employed with tacit geoscientific expertise~\citep{Kuckreja_2024_CVPR, zhan2024skyeyegpt, singh2024evalrs, guo2024remote, Singh_2024_CVPR}. Consider a geoscientist: if they use surface land temperature (SLT) products or electro-optical (EO) imagery under clear-sky conditions, they might substitute these with ground-station temperature data or synthetic aperture radar (SAR) images when cloud coverage exceeds a threshold. Unfortunately, encoding such nuanced ``SAR-over-EO'' logic into existing geospatial Copilots requires cumbersome prompting~\citep{stamoulis2025isrgpt}. Monolithic large language models (LLMs) become a bottleneck for achieving meaningful spatiotemporal scale in real-world RS applications due to their finite context windows and token capacities~\cite{hong2023metagpt}. Multi-agent Copilots have emerged as a novel paradigm, allowing agents to be added to or removed from a team of experts \textit{without} extra prompt tuning, enabling extensibility and simplifying development across numerous tasks~\cite{fourney2024magentic}. 

This work aims to harness the untapped potential of multi-agency in RS workflows. We present GeoLLM-Squad, a multi-agent system that \textit{separates agentic orchestration and geospatial task-solving}, capable of scaling to various RS tasks, including urban analysis, forestry, agriculture, and climate studies. Built on the open-source AutoGen~\citep{wu2023autogen} and GeoLLM-Engine~\citep{Singh_2024_CVPR} frameworks, we conduct comprehensive assessment of recent advancements in multi-agent orchestration~\cite{fourney2024magentic, lu2024chameleon} and workflow-based reasoning~\citep{fore2024geckopt, zhou2024webarena, paramanayakam2024less, wang2024agent} against human-expert curated workflows. Through this study, we identify key limitations of existing methods in geospatial contexts and propose a hybrid multi-agent approach that outperforms state-of-the-art baselines by up to 17\% in agentic correctness.

\vspace{-7pt}
\begin{table}[h]
\centering
\caption{Overview of recent geospatial agent paradigms.}
\label{tab:sota}
\scalebox{.9}{
\begin{tabular}{c c c c}
\toprule
\textbf{Tool usage} & \textbf{Agency} & \textbf{Application} & \textbf{Methods} \\ \cmidrule(lr){1-1} \cmidrule(lr){2-2}\cmidrule(lr){3-3}\cmidrule(lr){4-4}
\multirow{2}{*}{Single-turn/tool} & \multirow{2}{*}{Single-agent} & Satellite Vision, Urban & \citep{dias2024oreole, mall2024remote, hu2023rsgpt, silva2024large, Liu2024remoteclip, Zhang2024EarthGPT, jian2023stable}, \cite{bhandari2024urban, yu2024harnessing} \\ 
& & Map, Agriculture, Forest & \citep{zhang2024context}, \citep{li2024metafruit, yang2024multimodal, microsoft2023agriculture, bountos2023fomo, lacoste2024geo, zhu2024foundations, xiong2024neural} \\ \cmidrule(lr){1-3}
\multirow{2}{*}{Multi-turn/tool} & \multirow{2}{*}{Single-agent}  & Satellite Vision & ~\citep{Kuckreja_2024_CVPR, zhan2024skyeyegpt, singh2024evalrs, guo2024remote, Singh_2024_CVPR} \\
&  & DataOps, Climate & ~\citep{globeFlowGPT2024, chen2024llm}, \citep{goecks2023disasterresponsegpt} \\ \cmidrule(lr){1-4}
\multirow{3}{*}{Multi-turn/tool} & \multirow{3}{*}{Multi-agent}  & Satellite Vision, Map & \multirow{3}{*}{\textbf{GeoLLM-Squad}} \\
 &   & DataOps, Agriculture, & \\
 &   & Climate, Urban, Forest & \\
\bottomrule
\end{tabular}
}
\end{table}
\vspace{-7pt}

\section{Background and Novelty}

There is rapid progress in applying geospatial foundation models (GFMs) to RS tasks (Table~\ref{tab:sota}). Early approaches follow a single-turn-single-agent paradigm~\citep{hu2023rsgpt, Zhang2024EarthGPT}, where a fine-tuned LLM performs a narrowly defined task based on predetermined workflows. Recent advancements~\citep{Singh_2024_CVPR, chen2024llm} have introduced multi-turn-single-agent schemes that employ multi-tool APIs to support more complex, long-horizon tasks. However, these systems rely on static workflows (e.g., load-detect-plot ``chains'' in satellite vision)~\cite{Singh_2024_CVPR}. More critically, by centralizing all prompting, decision-making, reasoning, and tool execution within a single agent, these methods become bottlenecks when scaling to additional RS tasks~\cite{stamoulis2025isrgpt}.

\begin{figure*}[ht]
\centering
\includegraphics[width=0.98\linewidth]{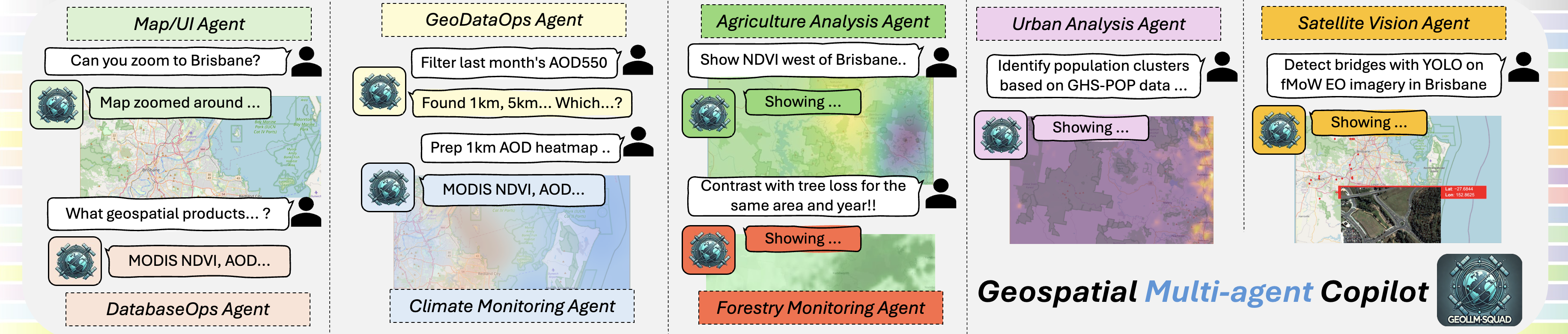}
\vspace{-5pt}
\caption{GeoLLM-Squad is a \textbf{\textit{multi-agent}} Copilot for RS workflows that builds on the open-source AutoGen~\citep{wu2023autogen} and GeoLLM-Engine~\citep{Singh_2024_CVPR} frameworks. Our orchestrator supports various geospatial agents and API tools to handle diverse workflows such as climate analysis or urban planning tasks.}
\label{fig:squad_overview}
\end{figure*}

\begin{table*}[ht]
\centering
\vspace*{-7pt}
\caption{Overview of geospatial products, metrics, and datasets used in GeoLLM-Squad workflows.}
\label{tab:datasets}
\scalebox{.95}{
\begin{tabular}{c c c c c}
    \toprule
    \textbf{Application} & \textbf{Metric} & \textbf{Dataset/Product} & \textbf{Resolution/Dates} & \textbf{User prompt examples (shortened for illustration)} \\ \cmidrule(lr){1-1} \cmidrule(lr){2-2} \cmidrule(lr){3-3} \cmidrule(lr){4-4} \cmidrule(lr){5-5}
    \colorbox{agrcolor}{Agriculture} & NDVI, Ref B2 & MOD13A3~\citep{Didan2015ndvi}, MOD09GA~\cite{vermote2015reflectance} & 1km, 2024  & \textit{From NDVI, recommend crop rotation areas in Brisbane} \\  \cmidrule(lr){2-5}
    \colorbox{clicolor}{Climate} & LST, AOD550 & MYD11A2~\citep{wan2021lst}, MCD19A2~\citep{Lyapustin2018aod} & 1km, 2024 & \textit{From LST, identify dangerous heatwave regions in Bundaberg} \\ \cmidrule(lr){2-5}
    \colorbox{urbcolor}{Urban} & Built-S, Popul. & GHS-BUILT-S/POP~\citep{GHSL2023data, pesaresi2023ghs} & 3 arcsec, 2020 & \textit{From 2020 pop., report overpopulation hotspots for Gympie} \\  \cmidrule(lr){2-5}
    \colorbox{forcolor}{Forest}  & Canopy, Treeloss & Tree Cover/Loss GFC~\citep{hansen2013forestchange} & 1 arcsec, 2020 & \textit{From 2020 canopy, recommend reforestation areas in Ipswich} \\\cmidrule(lr){2-5}
    \colorbox{satcolor}{Vision} & Detection, LCC & xView, FAIR, fMoW, ben~\citep{lam2018xview, gupta2019xview2, paolo2022xview3, luckett2024sarfish, sun2022fair1m, christie2018fMoW, sumbul2019bigearthnet}  & -- & \textit{From xView1, detect and plot aiplanes in Sydney} \\
    \bottomrule
\end{tabular}
}
\end{table*}

Multi-agent orchestration, such as composition-based~\citep{lu2024chameleon} and ledger-based scheduling~\citep{fourney2024magentic}, represents a novel agentic paradigm. Through comprehensive experimentation, we identify critical shortcomings in existing approaches when applied to RS workflows: ledger orchestration relies on frequent scheduling GPT calls~\cite{hong2023metagpt}, which becomes prohibitively expensive for RS tasks that involve loading, processing, analyzing, and plotting geospatial data. In contrast, while ledger-free compositional prompting eliminates this overhead, it struggles with error recovery~\citep{Singh_2024_CVPR}. To address these limitations, GeoLLM-Squad introduces a hybrid approach that combines composition-based reasoning with the iterative ledger-reassessment capabilities, as a ``best-of-both-worlds'' solution.

\section{Methodology}

\textbf{Agentic Environment.} We implement GeoLLM-Squad (Figure~\ref{fig:squad_overview}) on top of two components: GeoLLM-Engine~\citep{Singh_2024_CVPR} as the agentic frontend and AutoGen as the model serving backend~\citep{wu2023autogen}. The frontend integrates interactive map UIs with conversational ChatGPT-like functionality and API tools. The backend enables LLM function-calling for multi-agent communication and orchestration schemes~\citep{fourney2024magentic, lu2024chameleon}.

\textbf{Geospatial tasks}. We consider five RS workflows (Table~\ref{tab:datasets}) related to urban, agriculture, forest, climate, and satellite-vision studies. For forest data, we utilized tree canopy cover and loss from the Global Forest Change dataset~\citep{hansen2013forestchange}. For urban, we used the (Sentinel-2 and Landsat) GHS-BUILT-S and GHS-POP datasets from the Global Human Settlement layer~\citep{GHSL2023data, pesaresi2023ghs}. Climate workflows employed MODIS Terra products, namely 500m and 1km-resolution Aerosol Optical Depth (AOD 055) and land surface temperature (LST)~\citep{Lyapustin2018aod, wan2021lst}. Similarly, for agriculture data, we leveraged 1-Month/1km L3 Vegetation Indices and Band 2 surface reflectance~\citep{Didan2015ndvi, vermote2015reflectance}. MODIS products were downloaded from NASA Earthdata Search~\cite{nasa_earthdata_search} for the entire year 2024. Without loss of generality, our analysis covers the eastern Australian continent. For satellite vision tasks, we used xView1, xView2 (xBD), FAIR1M, xView3, SARFish, fMoW, and BigEarthNet~\citep{lam2018xview, gupta2019xview2, paolo2022xview3, luckett2024sarfish, sun2022fair1m, christie2018fMoW, sumbul2019bigearthnet}. All products (HDF, GeoTIFF) were standardized into GeoPandas formats for frontend-engine integration~\citep{Singh_2024_CVPR}.

\textbf{Agent tools}. We implement tools APIs to emulate functionalities from existing RS single-agents, such as~\cite{microsoft2023agriculture, bhandari2024urban, goecks2023disasterresponsegpt}, among others. The tools are tested and adapted against in-house workflows, i.e., by injecting status and error messages in their executable code for agentic use. We also repurpose map/UI, database, and data analytics functionalities from~\citep{Singh_2024_CVPR} as the Map, Database, and DataOps agents, respectively. Overall, GeoLLM-Squad comprise a total of 521 API functions, nearly a 3$\times$ increase compared to single-agent baselines~\cite{Singh_2024_CVPR, singh2024evalrs}.

\begin{table*}[ht!]
\centering
\caption{Agentic performance evaluation and downstream metrics across RS tasks with GPT-4o-mini~\cite{openai2025gpt4omini}.}
\label{tab:results_summary}
\scalebox{.9}{
\begin{tabular}{c c c S[table-format=2.2] S[table-format=2.2] S[table-format=2.2] S[table-format=2.2] S[table-format=2.2] S[table-format=2.2] S[table-format=2.2] S[table-format=2.2] S[table-format=2.2] S[table-format=2.2] S[table-format=2.2] S[table-format=2.2]}
\toprule
\multirow{4}{*}{\textbf{Method}} & \multicolumn{2}{c}{\textbf{Prompting}}  & \multicolumn{2}{c}{\textbf{Performance}} &
\multicolumn{2}{c}{\textbf{\colorbox{agrcolor}{Agriculture}}} & 
\multicolumn{2}{c}{\textbf{\colorbox{clicolor}{Climate}}} & 
\multicolumn{2}{c}{\textbf{\colorbox{urbcolor}{Urban}}} & 
\multicolumn{2}{c}{\textbf{\colorbox{forcolor}{Forestry}}} & 
\multicolumn{2}{c}{\textbf{\colorbox{satcolor}{Vision}}} \\ 
\cmidrule(lr){2-3} \cmidrule(lr){4-5} \cmidrule(lr){6-7} \cmidrule(lr){8-9} \cmidrule(lr){10-11}  \cmidrule(lr){12-13} \cmidrule(lr){14-15}
& TS & WM & \textbf{Crct.} & \textbf{Avg} & \textbf{NDVI} & \textbf{Ref B2} & \textbf{AOD550} & \textbf{LST} & \textbf{Built-S} & \textbf{Popul.} & \textbf{Loss } & \textbf{Canopy} & \textbf{LCC } & \textbf{Det } \\
& \citep{paramanayakam2024less} & \citep{wang2024agent} & \textbf{Rt\%}  & \textbf{Tokens} &  \textbf{ $\epsilon_{vi}\%$} & \textbf{ $\epsilon_{b2}\%$} & \textbf{ $\epsilon_{aod}\%$} & \textbf{ $\epsilon_{lst}\%$} & \textbf{ $\epsilon_{srf}\%$} & \textbf{ $\epsilon_{pop}\%$} & \textbf{ $\epsilon_{ls}\%$} & \textbf{ $\epsilon_{cvr}\%$} & \textbf{ $acc\%$} & \textbf{ $F1\%$} \\ 
\cmidrule(lr){1-1} \cmidrule(lr){2-2} \cmidrule(lr){3-3} \cmidrule(lr){4-4} \cmidrule(lr){5-5} \cmidrule(lr){6-6} \cmidrule(lr){7-7} \cmidrule(lr){8-8} \cmidrule(lr){9-9} \cmidrule(lr){10-10} \cmidrule(lr){11-11} \cmidrule(lr){12-12} \cmidrule(lr){13-13} \cmidrule(lr){14-14} \cmidrule(lr){15-15}
\multirow{3}{*}{GeoLLM-Eng.$^+$~\citep{Singh_2024_CVPR}} & -- & -- &  39.84  &    19.09  &  5.37    &  8.38    &  6.28  &  6.61    &  7.65    &  6.62    &  9.74    &  6.66    & 70.13    & 66.09  \\ 
 & \checkmark & -- &  41.86  &    20.09  &  7.74    & 12.49    &  8.62    &  9.43    & 10.70    & 10.73    &  9.00  &  7.20    & 86.25    & 70.53  \\
 & \checkmark & \checkmark &  43.32  &    22.16  &  5.35    &  8.52    &  4.06    &  5.27    &  4.36    &  5.55    &  5.90    &  3.89    & 81.88    & 64.16  \\ \cmidrule(lr){2-15}
\multirow{3}{*}{Chameleon~\citep{lu2024chameleon}} & -- & -- &  39.26  &    23.80  &  \textbf{4.77}    &  7.87    &  4.58    &  5.07    &  4.97    &  5.05    &  4.98    &  3.69    &  30.52    &  49.34  \\
 & \checkmark & -- &  40.14  &    21.49  &  4.87    &  8.13    &  4.32    &  4.78    &  4.24    &  5.03    &  5.21    &  4.03    & 33.89    & 57.97  \\
 & \checkmark & \checkmark &  41.03  &    58.40  &  4.66    &  7.66    &  \textbf{4.02}    &  5.00    &  4.98    &  \textbf{4.87}    &  5.14    &  4.08    & \textbf{85.82}    & 61.45  \\ \cmidrule(lr){2-15}
\multirow{3}{*}{Magentic~\citep{fourney2024magentic}}  & -- & - &  30.08  &   141.72  &  5.79    &  7.69    &  5.56    &  5.89    &  4.87    &  6.26    &  6.21    &  4.62    & 71.14    & 69.97  \\ 
 & \checkmark & -- &  30.37  &   188.07  &  5.31    &  7.87    &  5.13    &  5.33    &  \textbf{4.24}    &  5.80    &  6.48    &  4.30    & 54.18    & 73.20  \\ 
 & \checkmark & \checkmark &  33.98  &   210.11  &  7.67    & 10.15    &  6.36    &  7.62    &  8.05    &  7.22    &  7.67    &  5.41    & 73.08    & 76.59  \\\cmidrule(lr){2-15}
\textbf{GeoLLM-Squad} & \checkmark & \checkmark &  \textbf{60.29}  &    78.49  &  \textbf{4.77}    &  7.64    &  4.56    &  \textbf{4.14}    &  4.37    &  5.45    &  \textbf{4.58}    &  \textbf{3.61}    & 81.37    & \textbf{78.58}  \\
\bottomrule
\end{tabular}
}
\end{table*}

\textbf{Task generation}. Following~\citep{zhou2024webarena}, we provide the latest GPT-4o~\cite{hurst2024gpt} with user-task ``seeds'' and generate 56 sample prompts (7 per agent). We then construct human-curated solutions outlining the steps required to complete each sample. Next, we augment an ``oracle'' GPT-4o with these annotated solutions via RAG~\cite{wang2024agent}, prompting it to generate ``noisy'' ground-truths, which are then inspected by human annotators and the correct solutions are recorded as ``\textit{gold}'' ground-truths. We note that this process is the only offline step necessitating human annotation. We then use the human-vetted solutions and via few-shot ``training'' we prompt the oracle GPT-4o to generate 250 \textit{new} tasks (prompts) per agent along with their corresponding solutions, for a total of 2k tasks. This process results in representative ``\textit{pseudo}'' datasets that reflects realistic workflows as shown in~\cite{Singh_2024_CVPR, zhuang2023toolqa, narayan2024cookbook}. 

\textbf{Orchestrator}. Our scheduling follows compositional reasoning~\cite{lu2024chameleon} to devise program-like schedules in natural language (see below), which specify the sequential set of agents to execute the task and their (sub)prompts. Execution follows the prescribed schedule, and the Orchestrator aggregates return messages to check task completion. If incomplete, the schedule is revised and the loop is repeated; otherwise, the final response and the updated UI/map are returned to the user.

\textbf{Agents}. Each (sub)agent completes its assigned (sub)task using its dedicated toolkit via function calling. We note the advantage of having multiple experts; for example, the agriculture agent’s action space includes tools for crop rotation recommendations based on low-NDVI clusters without requiring to explicitly handle NDVI data loading.

\scalebox{.9}{
\begin{tcolorbox}[title=\textit{GeoLLM-Squad - Orchestrator prompting}, colback=gray!20, colframe=gray!75, rounded corners, sharp corners=northeast, sharp corners=southwest]
\textbf{$\gg$ Task}: You are a geospatial orchestrator [..]:

~~1. \textbf{Decompose} user request into agent subtasks.

~~2. \textbf{Generate} clear prompts for agents.

~~3. \textbf{Assign} agent order (e.g., load prior to filter).

\textbf{$\gg$ Example}: ``[..] 2024 NDVI data for Brisbane to [..] recommend crop rotation areas on the map.''

\textbf{$\gg$ Answer}: schedule = [\colorbox{dbcolor}{Database}(Load NDVI..), 
\colorbox{datcolor}{DataOps}(Filter Brisbane), 
\colorbox{agrcolor}{Agriculture}(Recommend crop rotation areas based on ..), 
\colorbox{mapcolor}{Map}(Plot..)]
\end{tcolorbox}
}

\scalebox{.9}{
\begin{tcolorbox}[title=\textit{GeoLLM-Squad - Agents prompting}, colback=gray!20, colframe=gray!75, rounded corners, sharp corners=northeast, sharp corners=southwest]
\textbf{$\gg$ Task}: You are a geospatial expert [..]. Solve [..]

\textbf{$\gg$ Query}: ``[..] 2024 NDVI data for Brisbane to [..] recommend crop rotation areas on the map.''

\textbf{$\gg$ TS}: \textit{Similar prompt}: ``crop areas for Syndey.''

\textit{Tools used}: \colorbox{agrcolor}{low\_ndvi\_clusters()}, [..] 

\textbf{$\gg$ WM}: \textit{Similar workflow}: ``NDVI [..] for Melbourne.''

\textit{Agents involved}: \colorbox{dbcolor}{Database}, \colorbox{datcolor}{DataOps}, [..] 
\end{tcolorbox}
}

To enhance agent prompting, we leverage two state-of-the-art practices, namely intent-based tool selection \textbf{TS}~\citep{paramanayakam2024less, fore2024geckopt} and agent workflow ``memory'' \textbf{WM}~\citep{wang2024agent, srinivasan2023nexusraven}. \textbf{TS} provides tool recommendations via similarity search against a precompiled ``training set'' of prompt-solution pairs, providing \textit{intra-agent} few-shot guidance. Orthogonally, \textbf{WM} compiles prompt-solution pairs at the workflow level, offering \textit{inter-agent} few-shot guidance. This improves agent reasoning and prompting. For instance, if the agriculture agent fails its task due to missing NDVI data, its response flags the dependency, prompting the Orchestrator to involve the database agent.

\begin{figure*}[hbt]
\centering
\includegraphics[width=0.98\linewidth]{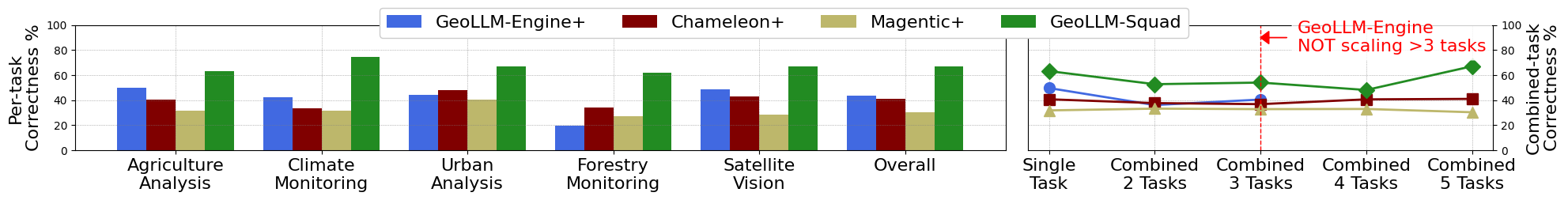}
\vspace{-10pt}
\caption{Single- \textit{vs}. multi-agent scalability ablations: we evaluate each method (with TS~\cite{paramanayakam2024less} and WM~\cite{wang2024agent} incorporated) separately for each task and for multi-task combinations. GeoLLM-Squad consistently provides better agentic performance across all tasks.}
\label{fig:results}
\end{figure*}

\section{Results}

\textbf{Metrics}. We follow established evaluation practices and we report agentic correctness and costs~\cite{fourney2024magentic, paramanayakam2024less}. Drawing from~\cite{zhou2024webarena,koh2024visualwebarena}, we compute \textit{correctness rate} as the ratio of correct tool-calling steps, i.e., how likely it is for the agent to invoke the correct functions in the expected order. To capture the impact of incorrect agentic actions to RS-specific metrics, we start from the fact that each task operates over sets of geospatial datapoints (e.g., NDVI values around location, or LST values for a date range). For each prompt, we record the data accessed by the agent and compare against the ``gold'' ground-truth set. For each missing item, we count its value as error and we compute mean-square percentage error $\epsilon$ following~\cite{cai2017neuralpower} across all 2k prompts. This assessment has been shown to better capture impact on downstream tasks instead of generic success rates or text-similary scores~\cite{singh2024evalrs, Singh_2024_CVPR}.

\textbf{Baselines}. Table~\ref{tab:results_summary} provides a comprehensive evaluation of agentic performance and metrics across different RS tasks with GPT-4o-mini~\cite{openai2025gpt4omini}. We compare GeoLLM-Squad against state-of-the-art methods, including the single-agent GeoLLM-Engine~\cite{Singh_2024_CVPR}, as well as the multi-agent Chameleon~\cite{lu2024chameleon} and Magentic~\cite{fourney2024magentic} frameworks. Note that, since GeoLLM-Engine fails to scale past 3 combined tasks (discussed next), we evaluate it on each individual task in isolation. 

GeoLLM-Engine correctness ranges from 39.84\% to 43.32\% for an average cost of 20.45k tokens, with the inclusion of tool selection (TS) and workflow memory (WM) improving performance. Magentic exhibits substantial communication overhead, with token costs exceeding 200k while achieving a correctness rate of only 33.98\%. Although it achieves competitive performance for some metrics, such as detection F1 score (76.59\%), the approach is inefficient due to excessive retry loops and repeated plan updates. Chameleon achieves efficient composition-based orchestration, reducing average token usage to about 40.14k with correctness rates comparable to the single-agent baseline. 

\textbf{GeoLLM-Squad}. By separating agentic orchestration from geospatial task-solving, GeoLLM-Squad significantly outperforms all baselines with a correctness rate of 60.29\%, reflecting a 17\% improvement compared to GeoLLM-Engine, while maintaining competitive token cost of 78.49k per task. Overall, we achieve the lowest error rates across most domains, including NDVI ($\epsilon_{vi}=4.77\%$), LST ($\epsilon_{lst}=4.14\%$), and tree-loss and coverage ($\epsilon_{ls}=4.58\%$, $\epsilon_{cvr}=3.61\%$), showcasing the ability to balance accuracy and agentic efficiency.

\textbf{Scaling across domains}. We evaluate each method separately for each task (denoted as $^+$). As shown in Figure~\ref{fig:results}, the per-task correctness across the considered baselines shows Chameleon outperforming GeoEngine-LLM in some tasks, while the single-agent performs better in others. Overall, GeoLLM-Squad consistently provides better correctness across all tasks. Next, we ablate the scalability limitations and we evaluate multi-task combinations for each method. As the number of combined tasks increases, global prompt complexity and token requirements grow significantly, leading to context-window constraints that cause single-agent systems to \textit{fail beyond three domains} (approximately 300 tools). In contrast, multi-agent setups like GeoLLM-Squad effectively distribute the toolset across specialized agents, maintaining robust performance as task complexity scales.

\textbf{Scaling across LLM families}. The performance of multi-agent systems depends on effective orchestration, a task where smaller language models (SLMs) typically struggle~\cite{hong2023metagpt}. For example, the Magentic framework explicitly emphasizes the necessity of GPT-based models for orchestration~\cite{fourney2024magentic}. However, this requirement imposes significant limitations for geoscientists due to cloud AI costs. Therefore, a critical property of effective multi-agent systems is their ability to perform well with open-source SLMs. To evaluate this, we test performance with Qwen-2.5 models (3B and 7B parameters)~\cite{yang2024qwen2} running on A100 GPUs (Table~\ref{tab:results_qwen}). Chameleon’s correctness rate drops by up to 20\% when replacing GPT-4o-mini with Qwen-2.5-3B, while Magentic collapses entirely to less than 9\% (effective ``noise'' levels) due to its dependence on advanced GPT models for its complex ledger-based logic. In contrast, GeoLLM-Squad demonstrates robust scalability across both SLMs, with the 7B variant achieving correctness rates comparable to the GPT-driven Chameleon. Of course, and despite GeoLLM-Squad achieving the highest performance among all methods, we emphasize the challenges faced by SLMs in performing agentic functions, as documented by practitioners in other domains~\cite{vllm2025}. We hope our findings motivate further \textit{advancements in open-source geospatial SLMs} to bridge this gap and enhance their applicability in multi-agent RS systems.

\begin{table}[t!]
\centering
\caption{Comparison across GPT-4o-mini and Qwen-2.5~\cite{yang2024qwen2} (7B, 3B) models. All methods incorporate TS~\cite{paramanayakam2024less} and WM~\cite{wang2024agent}}
\label{tab:results_qwen}
\begin{tabular}{c c S[table-format=2.2] S[table-format=2.2]}
\toprule
\textbf{LLM} & \textbf{Method} & \textbf{Crct. Rt \%} & \textbf{Avg $\epsilon \%$} \\
\cmidrule(lr){1-1} \cmidrule(lr){2-2} \cmidrule(lr){3-3} \cmidrule(lr){4-4}
\multirow{3}{*}{Qwen-2.5-7B} & Chameleon~\cite{lu2024chameleon} & 25.54 & 6.49 \\ 
~ & Magentic~\cite{fourney2024magentic} & 9.29 & \textit{--} \\    
~ & GeoLLM-Squad & \textbf{40.29} & \textbf{6.27} \\  \cmidrule(lr){1-1}
\multirow{3}{*}{Qwen-2.5-3B} & Chameleon~\cite{lu2024chameleon} & 21.11 & \textbf{6.52} \\ 
~ & Magentic~\cite{fourney2024magentic} & 7.81 & \textit{--} \\    
~ & GeoLLM-Squad & \textbf{36.95} & 7.31 \\ 
\bottomrule
\end{tabular}
\end{table}

\section{Future Work and Opportunities}

As the first work to apply multi-agency to RS workflows, to our knowledge, our findings highlight several opportunities for future research. First, as discussed, improving the agentic performance of SLMs remains an urgent challenge. This could involve injecting domain-specific RS knowledge into SLMs prompting workflows~\cite{stamoulis2025isrgpt} to compensate for the less powerful reasoning capabilities of SLMs. For the RS community, this is particularly important to ensure that tasks such as climate analysis and agricultural monitoring can be pursued with sustainable AI solutions~\cite{roscher2023data}, avoiding the prohibitive costs associated with multi-billion-parameter LLMs.

Second, while GeoLLM-Squad evaluates performance on multi-task benchmarks, these tasks primarily involve functional dependencies (e.g., simultaneously loading climate and agriculture data) rather than deeper semantic interdependencies. Future benchmarks should develop multi-agency workflows at a level beyond task delegation that require meaningful cross-domain reasoning~\cite{tsoumas2022evaluating}, such as enabling maritime analysts to correlate coastal drought conditions derived from climate data with spikes in illegal fishing activities~\cite{kroodsma2023global}. Current geospatial copilots lack the capability to handle this nuanced level of cross-agent synergy, presenting an exciting opportunity for the RS community.

\section{Conclusion}

Single-agent geospatial methods face scalability limitations when applied across multiple RS domains due to finite LLM context windows. In this work, we introduced GeoLLM-Squad, a multi-agent system that \textit{separates agentic orchestration and geospatial task-solving} processes which, combined with tool and workflow memory, improves both correctness and scalability across diverse RS applications. GeoLLM-Squad outperformed state-of-the-art baselines by up to 17\% in task correctness while maintaining competitive model efficiency.

\section*{Acknowledgements}

This work is partially supported by NSF CCF 2324854 and NSF-22-527. All open-source data products and frameworks used in this work, within non-commercial contexts and solely for research purposes, are fully cited in accordance with NASA ESDIS’s open data policy towards reproducibility. The views and conclusions expressed herein are those of the authors and do not represent official policies or endorsements of Microsoft, funding agencies, or the U.S. government.

\small
\bibliographystyle{IEEEtranN}
\bibliography{references}

\end{document}